\documentclass[11pt]{imsart}

\usepackage{graphicx}
\usepackage{amssymb}
\usepackage{amsmath}

\graphicspath{ {C:/cygwin64/home/Q10413/code/scripts/r/cog/2017/images/} }

\newtheorem{theorem}{Theorem}

\newtheorem{definition}[theorem]{Definition}
\newtheorem{remark}[theorem]{Remark}

\newcommand{\qed}{\nobreak \ifvmode \relax \else
      \ifdim\lastskip<1.75em \hskip-\lastskip
      \hskip1.75em plus0em minus0.75em \fi \nobreak
      \vrule height0.75em width0.50em depth0.25em\fi}

\startlocaldefs
\endlocaldefs

\begin{document}

\begin{frontmatter}

\title{A Brownian Motion Model and Extreme Belief Machine for Modeling Sensor Data Measurements}
\runtitle{EBM and Diffusions}

\author{\fnms{Robert A.} \snm{Murphy, 
Ph.D.}\ead[label=e1]{robert.a.murphy@wustl.edu}}
\address{\printead{e1}}

\runauthor{Murphy}

\begin{abstract}
As the title suggests, we will describe (and justify through the presentation of some of the relevant mathematics) prediction methodologies for sensor measurements.  This exposition will mainly be concerned with the mathematics related to modeling the sensor measurements.
\end{abstract}



\end{frontmatter}

\newpage

\tableofcontents

\newpage


\section{Prediction}
\label{prediction}

\subsection{Problem Statement}
\label{pproblem}

Given sensor data, devise a methodology to predict future sensor values.

\subsection{Solution Outline}
\label{psolution}

Suppose we have a measured slab of metal of known composition and we apply, to a specific spot on the slab, a heat source for a given duration of time and at a given temperature.  Then, not only does the specific spot see a rise in temperature, but other areas of the slab rise in temperature as well.   After the application of heat has ceased, we see a decrease in temperature across the entirety of the slab.  The dynamics of the spreading (and dissipation) of the heat are examples of diffusion processes.  Mathematically speaking, a \textit{diffusion process} is a sequence of dependent random quantities which can be traced back to some origin, with the randomness at each succeeding step being zero-mean Gaussian and having variance which only depends upon the elapsed time since the initial measurement.

For the processes from which the sensor data is measured, if it can be shown that the respective dynamics are diffusive, then the sensor measurements can be modeled as a distortion and translation of \textit{Brownian motion}.  Showing that the dynamics of the processes are diffusive is a matter of proving that the sensor measurements are zero-mean Gaussian, with variance that only depends upon the time since the initial measurement, along with each measurement taken since the initial time (future) being independent of measurements taken prior to the initial time (past).  The last statement is tantamount to the sensor data exhibiting the \textit{Markov property}.  Therefore, if it can be shown that a Markov process is an embedding in Brownian motion, then the sensor measurements can be modeled as a distortion and translation of a Markov process.  This is our approach for the model.\\

\noindent
1. Use a statistical test to show that the sequence of differences between sequential steps in the sensor measurements are normally distributed, resulting in the original sequence of sensor measurements exhibiting the Markov property, with the future measurements (beyond the initial measurement) being Gaussian.\\\\
2. With an inference of the data set being Gaussian, use maximum likelihood estimation to model the Markov process from which the data are sampled by sequentially averaging every two data points.  Pre\-pending the initial sample from the original data set, this new data set removes the noise, leaving the best (maximum likelihood) estimate of the underlying signal, while also retaining the Markov property, since each successive sample only depends upon the most recent one.  Furthermore, the most recent sample carries with it all information about previous samples, by design, while the initial sample remains the same as before.\\\\
3. Use Rogers $\cite[Section\ (I.5)\ \&\ Theorem\ (7.5)]{Rogers}$ to make the inference that the diffusion processes from which the data are sampled is just a distortion and translation of Brownian motion, resulting in the Markov process defined in (2) above.\\\\
4. Use Rogers $\cite[Theorem\ (6.1)]{Rogers}$ to conclude that the Markov process defined in (2) above is the distortion and translation of a linear combination of orthogonal functions having random, Gaussian-distributed Fourier coefficients.\\\\
5. If needed, estimate the orthogonal functions in the following manner, which when distorted, result in the Markov process from (2) above.  Consider the vector between each two successive data points in the sequence defined in (2) above.  Let $S$ be the number of vectors in total, which necessarily requires that $S+1$ is the number of data points in the sample.  For the current odd-indexed vector in the sequence of vectors, say $x_{k}$ for $k \in \{1,2,...,S-1\}$, there exists a constant $c_{k} \in (0,\infty)$ such that $c_{k}(x_{k}/\|x_{k}\|)$ is orthogonal to $x_{k+1}$ by defining
\begin{align*}
c_{k} &=
\begin{cases}
\|x_{k}\|+(y_{k}\|x_{k+1}\|)	& \text{if } \cos^{-1}(y_{k}) > \frac{\pi}{2} (\text{mod } \pi)\\
\|x_{k}\|	& \text{if } \cos^{-1}(y_{k}) = \frac{\pi}{2} (\text{mod } \pi)\\
\|x_{k}\|-(y_{k}\|x_{k+1}\|)	& \text{if } \cos^{-1}(y_{k}) < \frac{\pi}{2} (\text{mod } \pi)
\end{cases}
\end{align*}

\noindent
with $y_{k} = (x_{k+1} \cdot x_{k})/(\|x_{k+1}\|\|x_{k}\|)$ and $y_{k}\|x_{k+1}\|$ being the magnitude of the projection of $x_{k+1}$ onto the unit vector in the direction of $x_{k}$.  For $k \in \{1,2,...,S-1\}$, define
\begin{align*}
z_{k} &=
\begin{cases}
c_{k}(x_{k}/\|x_{k}\|)	& \text{if k is odd }\\
x_{k}	& \text{if k is even }
\end{cases}
\end{align*}
If $S$ is odd, then define $z_{S}=c_{S}(x_{S}/\|x_{S}\|)$, where
\begin{align*}
c_{S} &=
\begin{cases}
\|x_{S}\|+(y_{S}\|x_{S-1}\|)	& \text{if } \cos^{-1}(y_{S}) > \frac{\pi}{2} (\text{mod } \pi)\\
\|x_{S}\|	& \text{if } \cos^{-1}(y_{S}) = \frac{\pi}{2} (\text{mod } \pi)\\
\|x_{S}\|-(y_{S}\|x_{S-1}\|)	& \text{if } \cos^{-1}(y_{S}) < \frac{\pi}{2} (\text{mod } \pi)
\end{cases}
\end{align*}
where $y_{S} = (x_{S-1} \cdot x_{S})/(\|x_{S-1}\|\|x_{S}\|)$.  Otherwise, if $S$ is even, then $z_{S}=x_{S}$.  The equations of the lines through each of the orthogonal vectors (with appropriate domains) is the set of orthogonal functions that we seek.\\\\
6. For offline storage, retrieval and usage, generate models of the new data set found in (2) by using an \textit{Extreme Belief Machine (EBM)}.  An EBM is an \textit{Extreme Learning Machine (ELM)}, coupled with a \textit{Deep Belief Net (DBN)}, a class of energy-based, neural network model where each hidden layer is a \textit{restricted Boltzmann machine (RBM)} and the output layer is itself a single hidden-layer, feed-forward neural network with a sigmoid as its activation function (tantamount to a logistic regression at the output layer).\\\\
7. Learn \textit{association rules} for classification by first estimating $K$ the number of classes to form as the maximum of $2^{s}$ (where $s$ is the number of sensor features in the data set) and the estimated mean number of classes to form in Murphy $\cite{Murphy}$.  Next, in a B-tree classifier design, use random field theory to compute the energy of the RBMs from (6) used in the EBM that will successively divide the data until the level is reached where approximately $K$ classes are in the bottom level.  Note that in successively dividing each class at higher levels, we use the discontinuity (phase transition) in the EBM to demarcate each of the two new subclasses in the next division.  The $0-1$ combinations of sensor values, as well as the associated energy of the data in each leaf of the tree is an association rule.\\\\
8. Store the rules for retrieval and new classifications as rows of $0-1$ sensor values, an energy range for data in the class, and $0-1$ predictions.

\section{More on the Approach}

\subsection{Markov Property}
\label{prop}

Let $\{t_{k}\}_{k \ge 1} \subsetneq \mathbb{R}$ be a sequence of sampling times and $\{X_{t_{k}}\}_{k \ge 1}$ be a sequence of real-valued random variables.  If $X_{t_{k}}=x_{k}$ for all $k \ge 1$, then we know from Murphy $\cite{Murphy2}$ that if the error sequence $\{\epsilon_{k}=x_{t_{k+1}}-x_{t_{k}}\}_{k \ge 1}$ is normally distributed, the result is that the sequence of measurements exhibits the following property.

\begin{definition}
$\cite[Definition\ (1)]{Murphy2}$
The sequence $\{X_{t_{k}}\}_{k \ge 1}$ is said to exhibit the \textbf{Markov Property}, if there exists a measurable function $Y_{t_{1}}$ such that
\begin{equation}
\label{XY}
X_{t_{1}} = Y_{t_{1}}(X_{t_{0}})
\end{equation}
\end{definition}
for all sequential times $t_{0},t_{1} \in \mathbb{R}$ such that $t_{1} > t_{0}$.

Fix finite $K \ge 1$.  As in Murphy $\cite{Murphy2}$, we define a test statistic $W$ as
\begin{equation}
W = \frac{\bigg(\sum_{k=1}^{K}a_{k}x_{(k)}\bigg)^{2}}{\sum_{i=1}^{K}(x_{i}-\overline{x})^{2}},
\end{equation}
where $x_{(k)}$ and $\overline{x}$ are the $k^{th}$ element in an ordering of $\{x_{k}\}_{k = 1}^{K}$ and its sample mean, respectively, and $(a_{1},...,a_{K})$ is computed as
\begin{equation}
(a_{1},...,a_{K}) = \frac{m^{T}V^{-1}}{m^{T}V^{-1}V^{-1}m},
\end{equation}
such that $m = (m_{1},...,m_{K})$ is a vector of expected values of the order statistics used to give the ordering $\{x_{(k)}\}_{k=1}^{K}$ and $V$ is the covariance matrix of the order statistics.

\begin{definition}
\label{SW}
$\cite[Definition\ (8)]{Murphy2}$
The \textbf{Shapiro-Wilk Test of Normality} is the test statistic $W$, such that, if a level of significance ($p$-value) is assigned in a hypothesis test, where the null hypothesis is that the sample was drawn from a normal distribution, then a value of $W$ which exceeds the probability $(1-2p)$ affirms the null hypothesis.
\end{definition}

\noindent
The Shapiro-Wilk test now provides a sufficient condition for testing if the sequence of errors $\{\epsilon_{t_{k}}\}_{k = 1}^{K}$ is normally distributed.

\subsection{Maximum Likelihood Estimation (MLE)}
\label{mle}

It was shown in section $(\ref{prop})$ that the data exhibit the Markov property.  As before, suppose that $S+1$ is the number of samples and let $s_{k}$ be any sample in the data set for some $k \in \{1,2,...,S+1\}$.  Then, by the Markov property, $s_{k}$ only depends upon $s_{k-1}$.  In addition, $s_{k+1}$ is independent of $s_{k-1}$, given $s_{k}$, with the understanding that $s_{k-1}$ is a sample in the history prior to time $1$ whenever $k <= 1$ and $s_{k+1}$ is a sample in the future following time $S+1$ whenever $k >= S+1$.  As a result, the exhibition of the Markov property by the data lends itself to estimation of its signal by considering every two samples in sequence so that from section $(\ref{prop})$, the vector $x_{k+1}$ is defined by $x_{k},s_{k+1}$ and $s_{k+2}$ for $k \in \{1,2,...,S\}$, resulting in $S-1$ vectors.  We set $x_{1}=s_{1}$ in the new data set, in order to have the requisite $S$ vectors that model the distribution of the original data set.

Given that the data are normal, by results in section $(\ref{prop})$, we know from Hogg $\cite{Hogg}$ that for $k \in \{1,2,...,S\}$, the MLE of the set $\{s_{k+1},s_{k+2}\}$ is the average, $x_{k+1}=(s_{k}+s_{k+1})/2$.  By construction, the set of vectors $\{x_{k}\}_{k=1}^{S}$ exhibits the Markov property since each $x_{k+1}$ depends upon $x_{k}$ through common usage of $s_{k}$ in its computation, for all $k \in \{1,2,...,S-1\}$.  Furthermore, by this construction, the lines through each two consecutive vectors (with appropriate domains) gives rise to a piecewise continuous function that estimates the original distribution.  Finally, if we apply the orthogonalization process given in section $(\ref{psolution})$, then the lines through the orthogonal vectors (with appropriate domains) result in a set of basis functions for Brownian motion, such that the estimate of the distribution is recovered through a specific distortion and translation of the basis functions.

\subsection{Diffusions as Brownian Motion}
\label{dbm}

From section $(\ref{mle})$, we have used MLE to construct a Markov process that models the distribution of the original data set.  If we can embed the Markov process into a Brownian motion process, then the lines through the orthogonal vectors (with appropriate domains) constructed in the same section can be shown to form a set of basis functions for Brownian motion, and by embedding, distortion and translation, also a set of basis functions for the Markov process.  From Rogers $(\cite{Rogers})$, we have

\begin{definition}
Let $\Omega = \{\omega_{k}\}_{k=1}^{S+1}$ be the sample space from section $(\ref{psolution})$.  Then, $\mathcal{F}$ is a \textbf{$\sigma$-algebra} of subsets of $\Omega$, if\\
$(1)$ $\Omega \in \mathcal{F}$.\\
$(2)$ For any set $A \in \mathcal{F}$ such that $A \subseteq \Omega$, its complement, $A^{c} \subseteq \Omega$ is also an element of $\mathcal{F}$.\\
$(3)$ The union of any finite (or countably infinite) collection of elements of $\mathcal{F}$ is also an element of $\mathcal{F}$.
\end{definition}

\begin{definition}
A \textbf{probability space} is a triple $(\Omega,\mathcal{F},P)$, where $P$ is a probability distribution assigning values in $[0,1]$ to elements in $\mathcal{F}$.
\end{definition}

\begin{theorem}
\label{brown}
$\cite[Theorem\ (6.1)\ Brownian\ Motion]{Rogers}$
There exists a probability space upon which it is possible to define a process $(B_{t})_{t \in [0,1]}$ with the following properties\\
$(1)$ $B_{0}(\omega)=0$ for all $\omega \in \Omega$.\\
$(2)$ the map $t \mapsto B_{t}(\omega)$ is a continuous function of $t \in [0,1]$ for all $\omega \in \Omega$.\\
$(3)$ for every $0 \le s \le t \le 1$, $B_{t}-B_{s}$ is independent of $\{B_{u}\}_{u \le s}$ and has a $N(0,t-s)$ distribution.
\end{theorem}

By thm. $(\ref{brown})$, we know that Brownian motion exists and by prop's. $(2-3)$, it is a continuous process with the Markov property.  One might reason that any discrete set of samples are taken from a Brownian motion process whenever the samples exhibit the Markov property.  In this case, intuition leads to a correct conclusion, as the following result atests.

\begin{theorem}
\label{embed}
$\cite[Theorem\ (7.5)\ Skorokhod\ Embedding]{Rogers}$
Let $X$ be a zero-mean, random variable on $\Omega$ and let $E[X^{2}]=\sigma^{2}$ be the variance of $X$.  If $P$ is the distribution of $\Omega$, then there exists a Brownian motion process $(B_{t})_{t \in [0,1]}$ such that $P$ is its distribution.  Furthermore, if $T$ is a set of stopping times at which $P$ is sampled (using $X$) to produce $\Omega$, then $E[T]=\sigma^{2}$.
\end{theorem}

Now, by $\cite[Theorem\ (5.11)]{Rogers}$, we know that a diffusion is simply a specific distortion and translation of Brownian motion.  Hence, our Markov process can be distorted and translated to define a diffusion.  Lastly, from the proof of $\cite[Theorem\ (6.1)\ Brownian\ Motion]{Rogers}$, there exists orthogonal functions such that our Markov process is a linear combination of the orthogonal functions through the use of a specific set of Fourier coefficients which define the distortion and translation.  From section $(\ref{psolution})$, we know that the original Markov process is obtained from $\{z_{k}\}$ by multiplying each $z_{k}$ by $\|x_{k}\|/c_{k}$, for each $k \in \{1,2,...,S\}$.  Therefore, our Fourier coefficients used in the distortion and translation are given by $\|x_{k}\|/c_{k}$, for each $k \in \{1,2,...,S\}$.

\begin{remark}
For each $k \in \{1,2,...,S\}$, if $\cos^{-1}(y_{k}) = \pi/2$, then there is no distortion and translation of the $k^{th}$ orthogonal vector.  However, for each $k \in \{1,2,...,S-1\}$, if $\cos^{-1}(y_{k}) \ne \pi/2$, note that
\begin{eqnarray}
\label{dt}
\frac{\|x_{k}\|}{c_{k}} &=& \frac{\|x_{k}\|}{\|x_{k}\| \pm (y_{k}\|x_{k+1}\|)} \nonumber \\ &=& \frac{\|x_{k}\| \pm (y_{k}\|x_{k+1}\|) \mp (y_{k}\|x_{k+1}\|)}{\|x_{k}\| \pm (y_{k}\|x_{k+1}\|)} \nonumber \\ &=& 1 \mp \frac{y_{k}\|x_{k+1}\|}{\|x_{k}\| \pm (y_{k}\|x_{k+1}\|)}
\end{eqnarray}
so that by eqn. $(\ref{dt})$, it is clear how the $k^{th}$ orthogonal vector is distorted and translated to obtain $x_{k}$ as
\begin{eqnarray}
\label{distk}
x_{k} &=& z_{k}\frac{\|x_{k}\|}{c_{k}} \nonumber \\ &=& z_{k} \mp \frac{z_{k}(y_{k}\|x_{k+1}\|)}{\|x_{k}\| \pm (y_{k}\|x_{k+1}\|)}.
\end{eqnarray}
Similarly, if $k = S$ with $\cos^{-1}(y_{k}) \ne \pi/2$, then
\begin{eqnarray}
\label{dists}
x_{S} &=& z_{S}\frac{\|x_{S}\|}{c_{S}} \nonumber \\ &=& z_{S} \mp \frac{z_{S}(y_{S}\|x_{S-1}\|)}{\|x_{S-1}\| \pm (y_{S}\|x_{S-1}\|)}.
\end{eqnarray}
\end{remark}

\subsection{Extreme Belief Machine (EBM)}

\subsubsection{The Distribution}
\label{dist}

From section $(\ref{dbm})$, we know how to obtain $\{x_{k}\}_{k=1}^{S}$ from eqns. $(\ref{distk}-\ref{dists})$ as the distortions and translations of the orthogonal vectors $\{z_{k}\}_{k=1}^{S}$.  From Guyon $\cite[Sections\ (2.1.1-2.1.3)]{Guyon}$, our Markov process is one example of a (non-equilibrium) distribution of data which is derived from some common, unique equilibrium distribution.  Pairwise, sequential orthogonality of $\{z_{k}\}_{k=1}^{S}$ removes the additive noise from the vectors $\{x_{k}\}_{k=1}^{S}$, so that the orthogonal vectors define one such equilibrium distribution from which the data are derived.  Uniqueness (see Rogers $\cite[Section\ (I.6),\ pg.\ (13)]{Rogers})$ requires it to define the only such distribution, with probability $1$.  Hence, we can predict the next values from the Markov process as distortions and translations of predictions from the orthogonal set.

Let $\overline{X}$ be a random variable which takes as values, $\overline{x}=(s_{1},s_{2},...,s_{S+1})$, and let $p(\overline{x})$ be the probability density of $\overline{x} \in \mathbf{R}^{S+1}$.  From Guyon $\cite[Sections\ (2.1.1-2.1.3)]{Guyon}$,

\begin{equation}
p(\overline{x}) = P(\overline{X}=\overline{x}) = \frac{e^{H(\overline{x})}}{Z}
\end{equation}
is the density of the equilibrium distribution from the preceding paragraph, where $H$ is a linear function of the vector of data, $\overline{x} \in \mathbf{R}^{S+1}$, assigning an energy value to $\overline{x}$ and

\begin{equation}
Z = \sum_{\overline{x} \in \mathbf{R}^{S+1}} e^{H(\overline{x})}
\end{equation}
is a normalization constant making $p$ into a probability density.

Let $\mathcal{L}(\overline{x}) = \log{p(\overline{x})} = H(\overline{x})-\log{Z}$.  Since $\log{Z}$ is just a constant, we may take any constant terms in $H(\overline{x})$ to be zero and define $\mathcal{L}(\overline{x}) = H(\overline{x})+a$, for some constant $a \in \mathbf{R}$.  Then, $\mathcal{L}(\overline{x})$ completely determines the distribution, $p$, by a continuous transformation of the data.

\subsubsection{DBN as a Model}

Simply stated, a \textit{DBN} is a class of energy-based, neural network model where each hidden layer is an RBM.  In turn, an \textit{energy-based model} is typified by each data point in the sample set having an assigned value (state) from some finite set of values such that a function (Hamiltonian) of these values describes the “energy” in the data set.  Then, the model of the distribution of the sample data is a function of the energy, along with some unknown parameters (weights).

As usual, the idea is to find the set of weights which leads to the best (general) description of the sample data while minimizing the noise, which shows itself as increased energy in the distribution, resulting in more disorder.  As such, we want to minimize the energy in the model of the distribution, resulting in decreased disorder.  In essence, we seek an equilibrium (mean) distribution.

Finally, an \textit{RBM} is itself a single-layered, energy-based neural network with $P$ (from section $(\ref{dist})$) forming the model of the distribution of the sample data.  The Hamiltonian, $H$ (also from section $(\ref{dist})$), is a function of the states, as well as the Boltzmann constant and a “temperature” parameter for conducting \textit{simulated annealing} to find a minimal-energy, equilibrium distribution from which the original states were sampled.

\subsubsection{ELM for Simulated Annealing}

\textit{Simulated annealing} is an experiment such that states are switched at random and the temperature parameter is lowered by a certain amount until the model reaches equilibrium as a consequence of energy minimization.  We seek a method for conducting simulated annealing which will lead to the best estimate of $H$and $Z$, both as functions of the minimal energy state of $\overline{x}$ and the unknown weights.

Recall (from section $(\ref{dist})$) that $p$ is completely determined by $\mathcal{L}(\overline{x})$, which is a linear function of $H$ and $Z$.  From Huang $\cite{Huang2}$, we can use an ELM for simulated annealing to find the minimal-energy distribution of the data.  Each layer of the ELM is tantamount to an RBM with random values for both $H$ and $Z$, where multiplying constants providing the weighted average of the layered estimates are found by least squares, pseudo-inverse.  Likewise, since $\mathcal{L}(\overline{x})$ is continuous, then by Huang $\cite[Theorems\ (2.1-2.2)]{Huang2}$, it is guaranteed that the ELM will provide an exact approximation of $\mathcal{L}(\overline{x})$ as the number of layers, $L$, grows large.

\subsubsection{The EBM}

\begin{definition}
An \textbf{EBM} is an ELM such that each of its hidden layers consists of an RBM which approximates $\mathcal{L}(\overline{x})$ for the density, $p$, associated with a DBN.
\end{definition}

\subsubsection{Phase Transitions for Classification}

By $\cite[Theorem\ (1.11)]{Grimmett2}$, the density $p$ has a unique discontinuity (phase transition) that demarcates a transition from low probability states of sensor and event configurations to high probability ones.  As such, since the density $p$ is completely determined by $\mathcal{L}(\overline{x})$, then there are states $\overline{x}_{k},\overline{x}_{k+1}$ for $k \in \{1,2,...,S\}$ such that the change in probability is greatest when passing between these two states.  Therefore, we can compute the energy associated with each data point to obtain an ordering of the states.  Then, we can compute the associated changes in the energy function to identify the greatest change.  This change highlights the successive division between the classes in a B-tree hierarchy, as each leaf defining a conditional distribution in the hierarchy has a unique discontinuity by arguments in Guyon $\cite[Chapter\ (2)]{Guyon}$ concerning the conditional specification defined by the leaves at each level.  Indeed, once all discontinuities are found, further division into more classes should cease.  Consequently, a \textit{choice} is one or more divisions whose boundaries are demarcated by a phase transition.  If there is a preponderance of evidence (correlated class members) in one of the divisions, then by Grimmett $\cite[Theorem\ (1.11)]{Grimmett2}$ there is only one possible choice to be made, with no further division.  However, if a preponderance of evidence does not exist for one choice over all others, then Grimmett $\cite[Theorem\ (1.11)]{Grimmett2}$ also guarantees that many divisions exist, each with varying levels of supporting evidence (class members).

\end{document}